# Question Answering with Deep Neural Networks for Semi-Structured Heterogeneous Genealogical Knowledge Graphs

Omri Suissa[a], Maayan Zhitomirsky-Geffet[a] and Avshalom Elmalech[a]
[a]*Department of Information Science, Bar Ilan University, omrishsu@gmail.com, Israel*

**Abstract.** With the rising popularity of user-generated genealogical family trees, new genealogical information systems have been developed. State-of-the-art natural question answering algorithms use deep neural network (DNN) architecture based on self-attention networks. However, some of these models use sequence-based inputs and are not suitable to work with graph-based structure, while graph-based DNN models rely on high levels of comprehensiveness of knowledge graphs that is nonexistent in the genealogical domain. Moreover, these supervised DNN models require training datasets that are absent in the genealogical domain. This study proposes an end-to-end approach for question answering using genealogical family trees by: 1) representing genealogical data as knowledge graphs, 2) converting them to texts, 3) combining them with unstructured texts, and 4) training a transformer-based question answering model. To evaluate the need for a dedicated approach, a comparison between the fine-tuned model (Uncle-BERT) trained on the auto-generated genealogical dataset and state-of-the-art question-answering models was performed. The findings indicate that there are significant differences between answering genealogical questions and open-domain questions. Moreover, the proposed methodology reduces complexity while increasing accuracy and may have practical implications for genealogical research and real-world projects, making genealogical data accessible to experts as well as the general public.

## 1. Introduction

The popularity of "personal heritage", user-generated genealogical family tree creation, has increased in recent years, driven by new digital services, such as online family tree sharing sites, family tree creation software, and even self-service DNA analysis by companies like Ancestry and My Heritage. These genealogical information systems allow users worldwide to create, upload and share their family tree in a semi-structured graph format named GEDCOM (GEnealogical Data COMmunication)[1]. Most genealogical information systems also provide natural search capabilities (a search engine) to find relatives and related family trees. While the user interface [4, 10, 56, 75, 103] and user interactions [45, 69] with genealogical information systems are well researched, to the best of our knowledge, there is no research on natural question-answering in the genealogical domain for genealogical information systems.

As humans, we are accustomed to asking questions and receiving answers from others. However, the standard search engines and information retrieval (IR) systems require users to find answers from a list of documents. For example, for the question "How many children does Kate Kaufman have?", the system will retrieve a list of documents containing the words "children" and "Kate Kaufman". Unlike search engines and IR systems, natural question answering algorithms aim to provide precise answers to specified questions

---
[1] https://www.gedcom.org/

[47]. Thus, if a user is searching a genealogical database for the family tree of Kate Kaufman[2], a built-in question answering system will not return a list of possible matches but will provide a short and precise answer to various natural language questions. For instance, for a question such as "Where was Kate's father born?", a genealogical question answering system will return the answer "Hesse, Germany". Genealogical centers and museums seek to create a unique and personal experience for visitors using chatbots [73] and even holographic projections of private or famous people [78]. Hence, one practical implication of such a genealogical question answering system can be posing natural questions to a museum holographic character, or even a holographic restoration of a person from a family tree. Imagine walking into a genealogical center and talking to your great-grandmother, asking her questions about your family history and heritage. The underlying technology for such a conversation (inter alia) is based on the ability to answer natural questions on the GEDCOM data of genealogical family trees. The current state-of-the-art method for solving such a task is based on deep neural networks (DNN).

DNN models for open-domain natural question answering achieved high accuracy in multiple studies [15, 102, 116, 118, 119, 120, 127]. Training DNN models for question answering requires a golden standard dataset constructed from questions, answers, and corresponding texts from which these answers can be extracted. An extensive golden standard dataset for the natural question answering task widely used for training such models is Stanford Question Answering Dataset (SQuAD) [90, 91]. However, in the field of genealogy, there are no standard training datasets of questions and answers similar to SQuAD.

Generating a genealogical training dataset for question answering DNN is challenging, since genealogical data constitutes a semi-structured heterogeneous graph. It contains a mix of a structured graph and unstructured texts with multiple nodes and edge types, where nodes may include structured data on a specific person node (e.g., person's birthplace), structured data on a specific family node (e.g., marriage date), relations between nodes, and unstructured text sequences (e.g., bio notes of a person). Such a mix of structured heterogeneous graph data and unstructured text sequences is not the type of input that state-of-the-art models, like BERT [22] and other sequence-based DNN models, are designed to work with.

Therefore, the main objective of the proposed study is to design and empirically validate an end-to-end pipeline and a novel methodology for question-answering DNN using graph-based genealogical family trees combined with unstructured texts.

The research questions addressed in this study are:
1. What is the effect of the training corpus domain (i.e., open-domain vs. genealogical data) and the consanguinity scope on the accuracy of neural network models in the genealogical question answering task?
2. How to traverse a genealogical data graph while preserving the meaning of the genealogical relationships and family roles?
3. What is the effect of the question type on the DNN models' accuracy in the genealogical question answering task?

The main contributions of the study are:
1. A new automated method for question answering dataset generation derived from family tree data, based on the knowledge graph representation of genealogical data and its automatic conversion into a free text;
2. A new graph traversal method for genealogical data;
3. A fine-tuned question answering DNN model for the genealogical domain, Uncle-BERT, based on BERT[3] [22] that outperforms state-of-the-art DNN models (trained for answering open-domain questions) for various question types.

## 2. Related work

This section covers related work in the fields relevant to this research: genealogical family trees, neural network architecture, and question answering using neural networks.

---

[2] https://dbs.anumuseum.org.il/skn/en/c6/e22164995/Personalities/Kaufman_Kate

[3] https://huggingface.co/bert-base-uncased

*2.1. Genealogical family trees*

Genealogical family trees have become popular in recent years. Both non-profit organizations and commercial companies allow users worldwide to upload and update their family tree online. For example, commercial enterprises like Ancestry and My Heritage collect over 100 million[4] and 48 million[5] family trees, respectively; FamilySearch is the largest non-profit online collection of family trees with over a billion[6] unique individuals worldwide. Family trees can be created from various sources, such as family trees uploaded by private users (UGC) [6], clinical reports and DNA records [20, 105], biographical register [64], and even books [27]. Family tree records contain valuable information about individuals and their genealogical relationships, information that is useful for historical research and preservation [46], population and migration research [84], and even medical research [124, 126]. The user-generated content family trees phenomena, also called "personal heritage", combines the study of the history of one's ancestors with local and social history [6]. Figure 1 illustrates the degrees of relationships between two people in the genealogical domain [12].

---

[4] https://support.ancestry.com/s/article/Searching-Public-Family-Trees
[5] https://www.myheritage.co.il/about-myheritage/
[6] https://www.familysearch.org/en/about

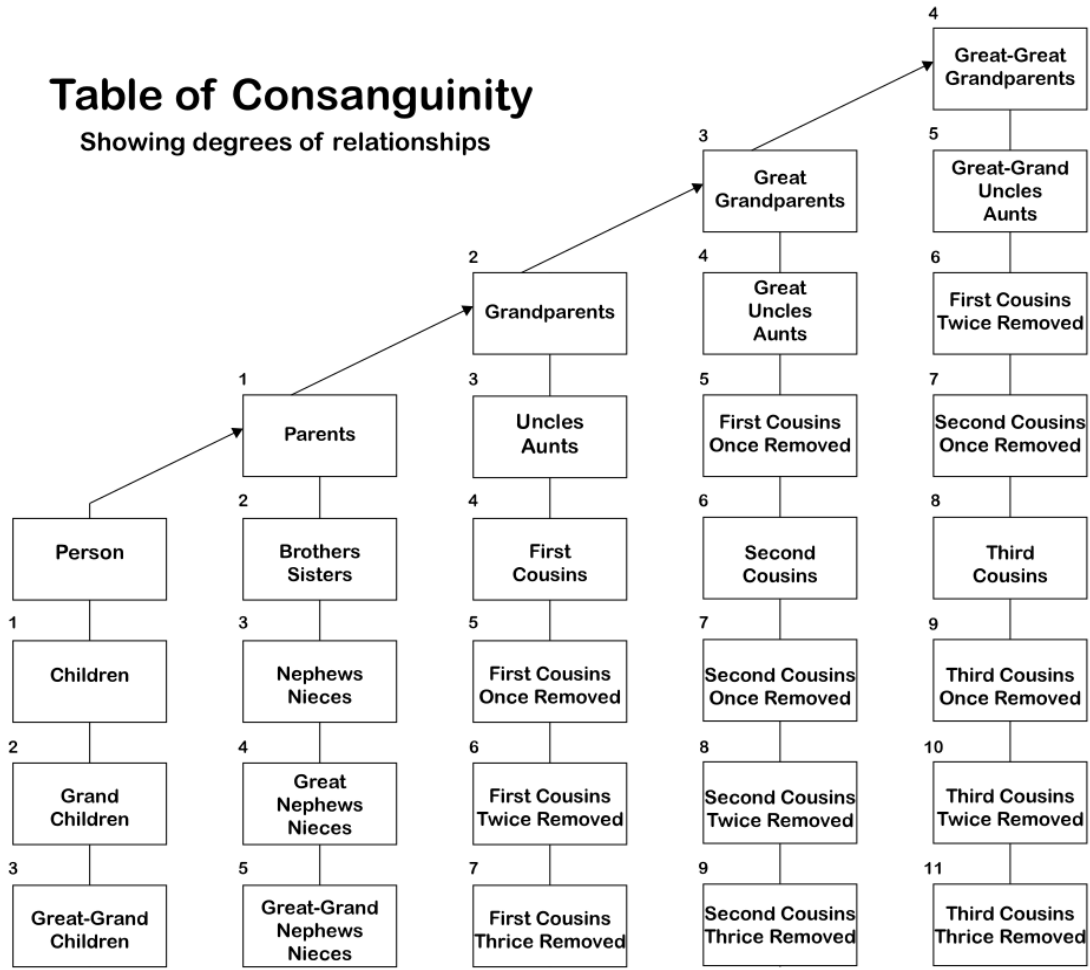

Fig. 1. Relation degrees in genealogy[7].

---

*2.1.1. The GEDCOM genealogical data standard*

The de facto standard in the field of genealogical family trees is the GEDCOM format [36, 56]. The standard developed by The Church of Jesus Christ of Latter-day Saints in 1984, and the latest released version (5.5.1) that was drafted in 1999 and fully released in 2019, still dominates the market [42]. Other standards have been suggested as replacements, but none were extensively adopted by the industry. GEDCOM is an open format with a simple lineage-linked structure, in which each record relates to either an individual or a family, and relevant information, such as names, events, places, relationships, and dates, appears in a hierarchical structure [36]. There are several open online GEDCOM databases, including GenealogyForum [35], WikiTree[8], GedcomIndex[9], Anu Museum[10], Ancestry.com, and others.

In GEDCOM format, every person (individual) in the family tree is represented as a node that may contain known attributes, such as first name, last name, birth date and place, death date and place, burial date and place, notes, occupation, and other information. Two individuals are not linked to one another directly. Each individual is linked to a family node as a "spouse" (i.e., a parent) or a "child" in the family. Figure 2 shows a sub-graph corresponding to a Source Person (SP) whose data is presented in the GEDCOM file in Figure 3. Each individual and family are assigned a unique ID – a number bracketed by @ symbols and a class name (INDI – individual, FAM – family). The source person is noted as SP (@I137@ INDI - Emily Williams in the GEDCOM file), families as F and other persons as P. In this example, P3, P4, P5, and P6 are the grandparents of SP; P1 and P2 are SP's parents in family F1 (@F79@ in the GEDCOM file); P7 and P8 are SP's siblings; P10 (@I162@ INDI – John Williams in the GEDCOM file) is SP's spouse from family F4 (@F73@ in the GEDCOM file), P12 and P13 are SP's children; and P15, P16, and P17 are SP's grandchildren. Moreover, as seen in Figure 3, SP was a female, born on 28 MAY 1816 in New York, USA, who died on 7 FEB 1899 in Uinta, Wyoming, USA, and was buried three days later in the same place. Furthermore, SP was baptized on 1 JUN 1832 and was endowed on 30 DEC 1845 in TEMP NAUVO (maybe Nauvoo Temple[11], Illinois). Her husband, P10, John Williams, was a male, born on 16 MAY 1826 in Indiana, USA, who died on 25 SEP 1912 in Uinta,

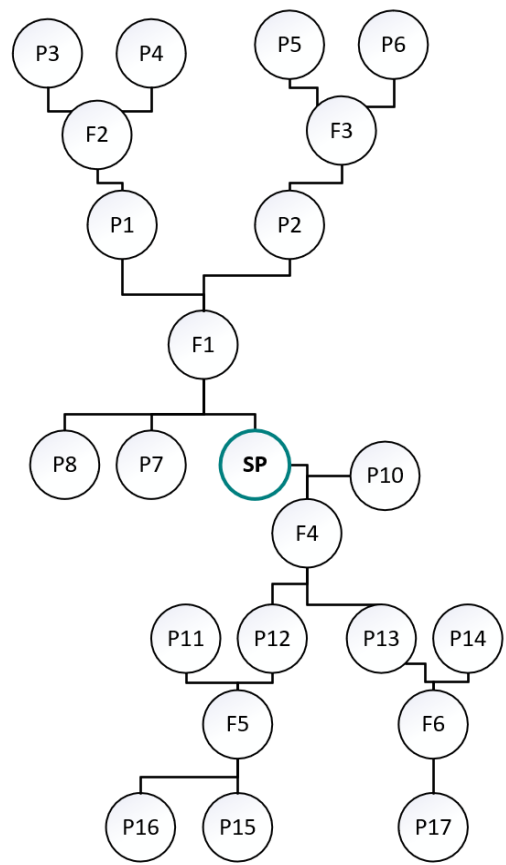

Fig. 2. Family tree structure.

Wyoming, USA, and was buried three days later in the same place. He was baptized on 9 AUG 1877, although there is a note stating that it may be on the 12 of AUG 1877, and he was endowed with his wife. For practical reasons, the GEDCOM file example in Figure 3 contains only a small portion of the data presented in Figure 2.

---

[8] https://www.wikitree.com/
[9] http://gedcomindex.com/gedcoms.html
[10] https://dbs.anumuseum.org.il
[11] https://churchofjesuschristtemples.org/nauvoo-temple/

```
0 HEAD
1 SOUR SomeSite
2 NAME Some Site
2 VERS 3.0
1 DATE 30 JUN 1985
2 TIME 19:38:50
1 FILE example.ged
1 GEDC
2 VERS 5.5
2 FORM LINEAGE-LINKED
...
1 CHAR ANSEL
0 @I137@ INDI
1 NAME Emily Williams
1 SEX F
1 BIRT
2 DATE 28 MAY 1816
2 PLAC New York, USA
1 DEAT
2 DATE 7 FEB 1899
2 PLAC Uinta, Wyoming, USA
1 BURI
2 DATE 10 FEB 1899
2 PLAC Uinta, Wyoming, USA
1 BAPL
2 DATE 1 JUN 1832
1 ENDL
2 DATE 30 DEC 1845
2 TEMP NAUVO
1 FAMS @F73@
1 FAMC @F79@
1 SLGC
2 DATE 18 NOV 1894
2 TEMP SLAKE
1 CHAN
2 DATE 14 MAY 1999
3 TIME 09:57:42
0 @I162@ INDI
1 NAME John Williams
1 SEX M
1 BIRT
2 DATE 16 MAY 1826
2 PLAC Indiana, USA
1 DEAT
2 DATE 25 SEP 1912
2 PLAC Uinta, Wyoming, USA
1 BURI
2 DATE 28 SEP 1912
2 PLAC Uinta, Wyoming
1 BAPL
2 DATE 9 AUG 1877
1 ENDL
2 DATE 30 DEC 1845
2 TEMP NAUVO
1 FAMS @F73@
1 FAMC @F1598@
1 NOTE Baptism date appears to be 3 days later by the records of the city...
...
```

Fig. 3. (part of) GEDCOM family tree file.

*2.2. Question answering using DNN*

A DNN is a computational mathematical model that consists of several "neurons" arranged in layers. Each neuron performs a computational operation and transmits the computed information (calculation result) to the neurons in the next layer. The information is passed over and changed from layer to layer until it becomes the output in the network's last layer. The conventional learning method is backpropagation, which refers to learning as an optimization problem [122]. After each training cycle, a comparison between the network prediction (output) and the actual expected result is performed, and a "loss" (i.e., the gap) is calculated to estimate the changes needed in the network operations (the weight of neuron's transformation). Changes in the network weights are usually performed using the Gradient Descent methods [7].

In recent years, DNNs have become the state-of-the-art method for text analysis in the cultural heritage space [110], and natural language question-answering systems based on DNN have become the state-of-the-art method for solving the question answering task [61]. The underlying task of question answering is Machine Reading Comprehension (MRC), which allows machines to read and comprehend a specified context passage for answering a question, similarly to language proficiency exams. Question answering, on the other hand, aims to answer a question without a specific context. These QA systems store a database containing a sizeable unstructured corpus and generate the context in real-time based on relevant text passages to the input question [138]. Due to the magnitude of comparisons needed between the query and each text passage in the corpus, and due to the number of calculations (a large number of multiplications of vectors and matrices) when a DNN model predicts the answer span for every given text passage, DNNs are not applied on the entire database of texts, but only on a limited number of passages. Hence, when a user asks a question, the system searches [12] the database for K passages that are relevant to the user question. The system will then use the DNN model to predict the answer span (start and end positions) for each text passage (from the K passages) with a confidence level. The answer with the highest confidence level is selected as the answer to be presented to the user. Thus, a typical pipeline (shown in Figure 4) of DNN for question answering will be a compound of (1) two inputs - (a) a text passage (i.e., a document) that may contain the answer, and (b) a question; and (2) two outputs: (a) the start index of the answer in the text passage, and (b) the end index of the answer in the text passage. The inputs are encoded into vectors using static embeddings methods, such as Word2Vec [77] and GloVe [86] or using contextualized embeddings of words like Bidirectional Encoder Representations from Transformers (BERT) [22], Embeddings from Language Models (ELMo) [87] and other methods [88]. One of the main advantages of contextual embeddings is the ability to handle disambiguations of words and entities [81, 129]. The input vector is transferred through the network, and the final layer output vectors are the probability of every word to be the start or the end of the span (i.e., answer). The score of every span is a combination of the start and end tokens' probabilities. The most probable span is then translated back to a sequence of words using the embedding method [22] (see section 3.2 for a more detailed description). Researchers proposed various DNN-based models to solve the task of finding (ranking) an answer span (the part of the text that contains the answer for the question) in the document [22, 97, 118, 119, 133] or a single sentence [34, 62].

---
[12] A common approach for finding relevant passages is reverse indexing [11, 53, 54, 71, 104]

### 2.2.1. Natural question answering using DNN architecture

Over the years, different deep learning layers have been developed with various abilities. Until recently, the typical architecture for natural language questions answering was based on Recurrent Neural Networks (RNN) such as Long Short Term Memory (LSTM) [48] and Gated Recurrent Units (GRU) layers [17]. RNN layers allow the network to "remember" previously calculated data and thus learn answers regarding an entire sequence. These layers are used to construct different models, including a sequence-to-sequence model [112] that uses an encoder-decoder architecture [17] that fits the question-answering task. This model maps a sequence input to a sequence output, like a document (sequence of words) and a question (sequence of words) to an answer (sequence of words) or to classify words (whatever the word is the start or the end of the answer). RNN architecture often works with direct and reverse order sequences (bidirectional-RNN) [96]. It may also include an attention mechanism [115], which "decides" (i.e., ranks) which parts in the sequence are more important than others during the transformation of a sequence from one layer to another.

Another typical architecture is based on a Convolutional Neural Network (CNN). Unlike RNNs, CNNs architecture does not have any memory state that accumulates the information from the sequence data. CNN architecture uses pre-trained static embeddings where each CNN channel aggregates information from the vectorial representation. Channels of different sizes enable it to deal with n-gram-like information in a sentence [57].

Question answering task can also be modeled as a graph task (e.g., traversal, subgraph extraction). The data can be represented as a knowledge graph (KGQA), where each node is an entity, and each edge is a relation between two entities. When answering the question, the algorithm finds the relevant entities for the question and traverses over the relations or uses the node's attributes to find the answer node or attribute [13, 24, 134]. To work with graphs, Graph Neural Networks (GNN) [94] models have been developed that operate directly on the graph structure. GNN can be used for resolving answers directly from a knowledge graph by predicting an answer node from question nodes (i.e., entities) [29, 38, 72, 80, 101, 134]. The GNN model is similar to RNN in the sense that it uses near nodes and relations (instead of previous and next token in RNN) to classify (i.e., label) each node. However, these models cannot directly work with unstructured or semi-structured data or rely on the ability to complete and update the knowledge graph from free texts using knowledge graph completion tasks, such as relation extraction [8, 82, 128] or link prediction [32, 52].

An improved approach considered to be the state-of-the-art in many NLP tasks, including question answering, is Transformers architecture [115], which uses the attention mechanism with feed-forward layers (not RNNs); this kind of attention is also called Self Attention Network (SAN). Well-known examples of SANs are Bidirectional Encoder Representations from Transformers (BERT) [22] and GPT-2 [89] models. Several BERT-based models were developed in recent years [125], achieving state-of-the-art performance (accuracy) in different question answering tasks. These include RoBERTa - a BERT model with hyperparameters and training data size tuning [70]; DistilBERT - a smaller, faster, and lighter version of BERT [93]; ELECTRA – a BERT-like model with a different training approach [18]. Although standard BERT-based models receive textual sequence as input, all the above architectures can also be mixed. For example, a Graph Convolutional Network (GCN) [114] can be utilized for text classification by modeling the text as a graph and using the filtering capabilities of a CNN [131].

There are several question-answering DNN pipelines based on knowledge graphs that support semi-structured data (a mix of a structured graph and unstructured texts) [29, 40, 134, 137]. As shown in Figure 5, a current state-of-the-art pipeline of this type,

Deciphering Entity Links from Free Text (DELFT) [134], uses the knowledge graph to extract related entities and sentences, filters possible textual sentences using BERT, and then traverses a filtered subgraph using a GNN. The pipeline starts with identifying the entities in the question. Then, related entities ("candidates") from the knowledge graph and relevant sentences ("evidence relations") from unstructured texts are extracted and filtered using BERT. A new subgraph is generated using the question entities, the filtered evidence relations, and the candidate entities. Using this subgraph, a GNN model learns to rank the most relevant node. Thus, the model obtains a "trail" from the question nodes to a possible candidate node (i.e., answer). The pipeline applies two DNN models: a BERT model to rank the evidence relations and a GNN model to traverse the graph (i.e., predict the answer node).

However, these methods, using the unstructured texts to create or complete the knowledge graph, rely heavily on well-defined semantics and fail to handle questions with entities completely outside the knowledge graph or questions that cannot be modeled within the knowledge graph. For example, Differentiable Neural Computer (DNC) [38] can be used to answer traversal questions ("Who is John's great-great-grandfather?"), but not to answer content-related questions when the answer is written in the person's bio notes (e.g., "When did John's great-great-grandfather move to Florida?"). As part of the evaluation experiments in this study, the performance of the above mentioned DELFT pipeline, adapted to the genealogical domain, was compared to that of the proposed pipeline.

In summary, the generic question answering pipelines described above cannot be applied as-is in the genealogical domain, without compromising on accuracy, for the following reasons: (1) The raw data is structured as graphs, each graph contains more information than a DNN model can handle in a single inference process (each node is equivalent to a document), (2) A user may ask about different nodes and different scopes of relations (i.e., different genealogical relation degrees); (3) There is a high number of nodes containing a relatively small volume of structured data and a relatively large volume of unstructured textual data. In addition, the vast amount of different training approaches, hyperparameters tuning, and architectures indicate the complexity of the models and sensitivity to a specific domain and sub-task.

The question answering approach proposed in this study simplifies the task pipeline by converting the genealogical knowledge graph into text, which is then combined with unstructured genealogical texts and processed by BERT's contextual embeddings. Converting the genealogical graph into text passages can be performed using knowledge-graph-to-text templates and methodologies [21, 26, 55, 76, 123], and knowledge-graph-to-text machine learning and DNN models [5, 33, 63, 66, 68, 78, 79, 99, 106]. Template-based knowledge-graph-to-text methods use hardcoded or extracted linguistic rules or templates to convert a subgraph into a sentence. Machine learning and DNN models can be trained to produce a text from knowledge-graph nodes. The input for a knowledge-graph-to-text model is a list of triples of two nodes and their relation, and the output is a text passage containing a natural language text with input nodes and their relations as syntactic sentences. To this end, DNN models are often trained using commonsense knowledge graphs of facts, such as ConceptNet [107], BabelNet [83], DBpedia [3], and Freebase [85], where nodes are entities, and the edges represent the semantic relationships between them. Some models use the fact that knowledge graphs are language-agnostic to generate texts in multi-languages (e.g., [79]).

*2.3. Questions and answers generation for DNN-based question answering systems*

Training of a DNN question answering model requires a set of text passages and corresponding pairs of questions and answers. Multiple approaches exist for generation of questions (and answers): knowledge-graph-to-question template-based methodology (similar to the context generation) [67, 98, 136, 140], WH questions (e.g., Where, Who, What, When, Why) rule-based approach [80], knowledge graph-based question generation [16, 50], and DNN-based models for generating additional types of questions [25, 49, 117, 135]. The rule-based method uses part-of-speech parsing of sentences using the Stanford Parser [59], creates a tree query language and tree manipulation [65], and applies a set of rules to simplify and transform the sentences to a question. To guarantee question quality, questions are ranked by a logistic regression model for question acceptability [44]. The DNN question generation models are trained on SQuAD [90, 91] or on facts from a knowledge graph to predict the question and its correct answer from the context (i.e., the opposite task from question answering) using bi-directional [96] LSTM [48] encoder-decoder [17] model with attention [115].

This study adopted the format of the SQuAD dataset, which is a well-known benchmark for machine learning models on question answering tasks with a formal leaderboard[13]. SQuAD is a reading comprehension dataset consisting of questions created by crowd workers on a set of Wikipedia articles. The answers to the questions are segments of text from the corresponding reading passage (context), or the question might be unanswerable. SQuAD 2.0 combines 100,000 questions and answers and over 50,000 unanswerable questions written adversarially by crowd workers to look similar to answerable ones. To do well on SQuAD 2.0, natural question answering models must answer questions when possible and determine when no answer is supported by the paragraph, in which case they must abstain from answering.

SQuAD 2.0 is a JSON formatted dataset, presented in Figure 6, where each topic (a Wikipedia article) has a *title* and *paragraphs*. Each paragraph contains a *context* (text passage) and questions (*qas*). Each question contains the *question* text, *id,* may contain *answers* (if it is answerable), may contain *plausible answers*, or be marked as *impossible*. Each answer is constructed from a *text* and a *start index* (the word index) of the answer in the text passage.

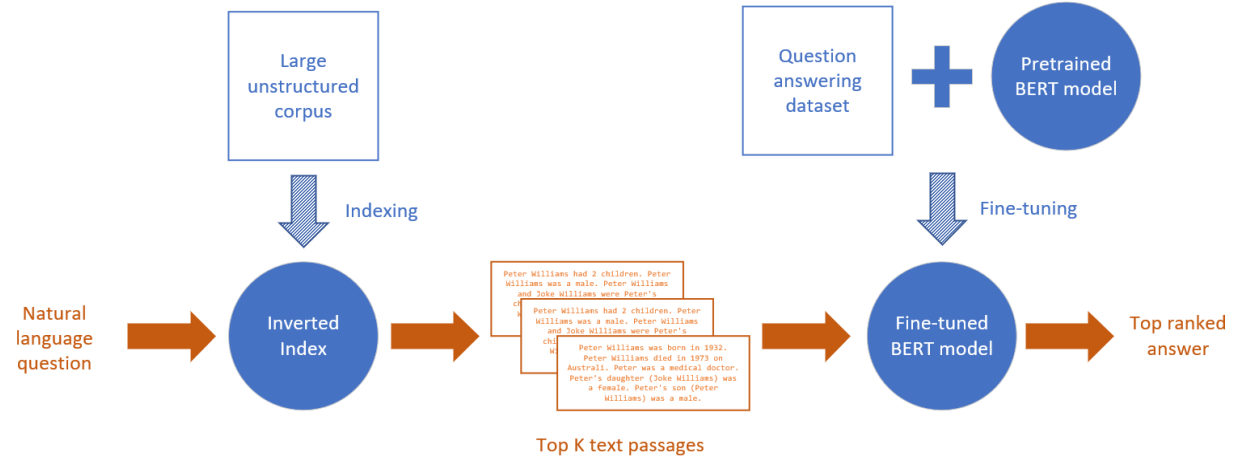

Fig. 4. Typical open-domain question answering pipeline.

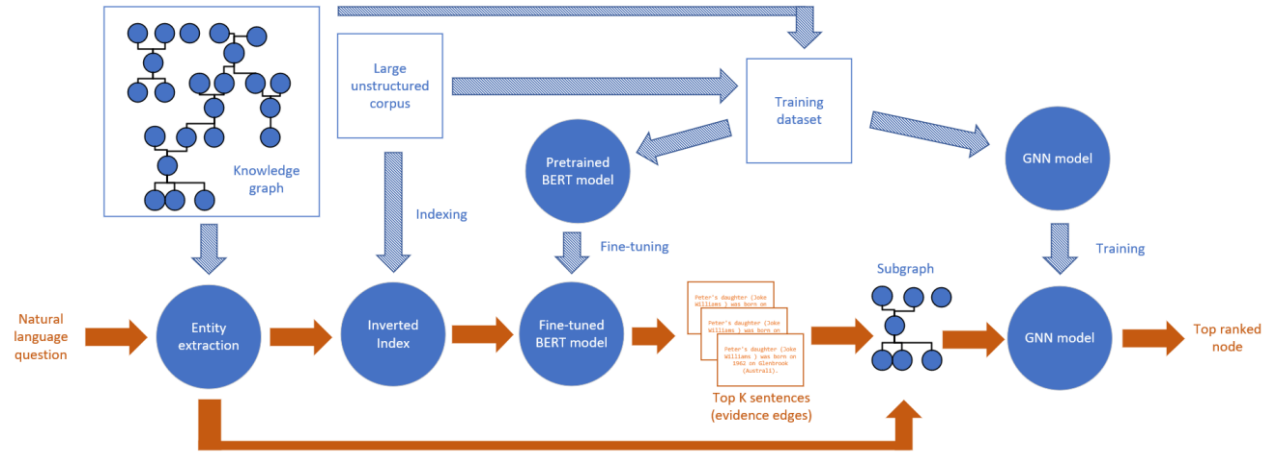

Fig. 5. Typical knowledge graph question answering pipeline.

---

[13] https://rajpurkar.github.io/SQuAD-explorer/

```json
{ "version": "v2.0",
  "data": [
   { "title": "Normans",
     "paragraphs": [
       { "qas": [
           { "question": "Who was the duke in the battle of Hastings?",
             "id": "56dddf4066d3e219004dad5f",
             "answers": [{ "text": "William the Conqueror", "answer_start": 1022}],
             "is_impossible": false },
           { "question": "Who ruled the duchy of Normandy",
             "id": "56dddf4066d3e219004dad60",
             "answers": [ { "text": "Richard I", "answer_start": 573 }],
             "is_impossible": false },
           { "plausible_answers": [
             { "text": "political, cultural and military",
               "answer_start": 31 }],
             "question": "What type of major impact did the Norman dynasty have on modern Europe?",
             "id": "5ad3a266604f3c001a3fea27",
             "answers": [],
             "is_impossible": true},
           … ],
         "context": "The Norman dynasty had a major political, cultural and military impact on medieval Europe and even the Near East. The Normans were famed for their martial spirit and eventually for their Christian piety, becoming exponents of the Catholic orthodoxy into which they assimilated. They adopted the Gallo-Romance language of the Frankish land they settled, their dialect becoming known as Norman, Normaund or Norman French, an important literary language. The Duchy of Normandy, which they formed by treaty with the French crown, was a great fief of medieval France, and under Richard I of Normandy was forged into a cohesive and formidable principality in feudal tenure. The Normans are noted both for their culture, such as their unique Romanesque architecture and musical traditions, and for their significant military accomplishments and innovations. Norman adventurers founded the Kingdom of Sicily under Roger II after conquering southern Italy on the Saracens and Byzantines, and an expedition on behalf of their duke, William the Conqueror, led to the Norman conquest of England at the Battle of Hastings in 1066..."
       },
       … ]
   },
   … ]
 }
```

Fig. 6. SQuAD 2.0 JSON format example.

## 3. Methodology

While using DNNs for the open-domain question answering task has become the state-of-the-art approach, automated question answering systems for genealogical data is still an underexplored field of research. This paper presents a new methodology for a DNN-based question answering pipeline for semi-structured heterogeneous genealogical knowledge graphs. First, a training corpus that captures both the structured and unstructured information in genealogical graphs is generated. Then, the generated corpus is used to train a DNN-based question answering model.

### 3.1. Gen-SQuAD generation and graph traversal

The first phase in the proposed methodology is to generate a training dataset using the text sequence encoding with a graph traversal algorithm. This dataset should contain questions with answers and free text passages from which the model can retrieve these answers.

Generating a training dataset from genealogical data is a three-step process. The result of the process is Gen-SQuAD, a SQuAD 2.0 format dataset tailored to the genealogical domain. As shown in Figure 7, the process includes the following steps: (1) decomposing the GEDCOM graphs to CIDOC-CRM-based [14] knowledge sub-graphs, (2) generating text passages from the obtained knowledge sub-graphs, and (3) generating questions and answers from the text passages. Finally, the context and matching questions and answers are saved in the SQuAD 2.0 JSON format. The following sections present in detail each step of the Gen-SQuAD generation process.

---
[14] http://www.cidoc-crm.org/

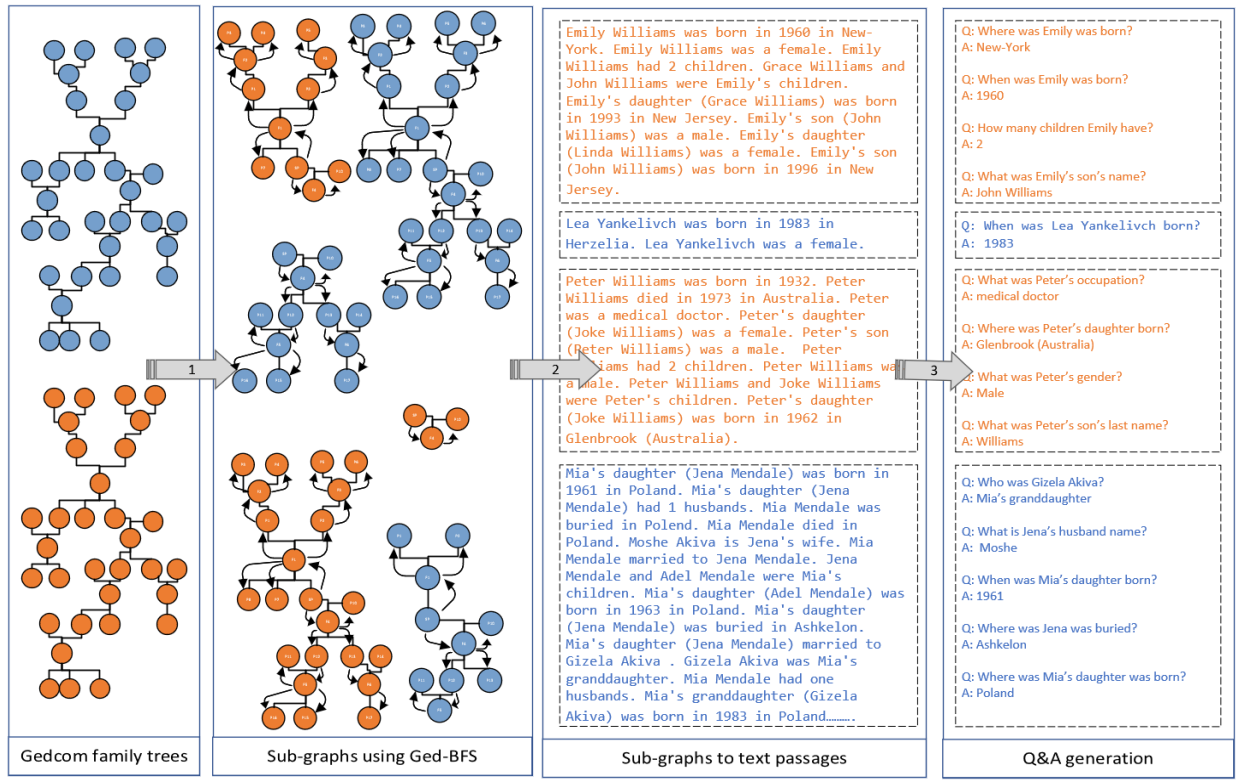

Fig. 7. Gen-SQuAD generation.

*3.1.1. Sub-graph extraction and semantic representation*

While there are some DNN models that can accept large inputs [9, 58], due to computational resource limitations, many DNN models tend to accept limited size inputs, usually ranging from 128 to 512 tokens (i.e., words) [141]. However, family trees tend to hold a lot of information, from names, places, and dates to free-text notes, life stories, and even manifests. Therefore, using the proposed methodology, it is not practical to build a model that will read an entire family tree as an input (sequence), and it is necessary to split the family tree into sub-trees (sub-graphs). Several generic graph traversal algorithms may be suitable for traversing a graph and extracting sub-graphs, such as Breadth-First-Search (BFS) and Depth-First-Search (DFS). BFS's scoping resembles a genealogy exploration process that treats first relations between individuals that are at the same depth level (relation degree) in the family tree, moving from the selected node's level to the outer level nodes. However, the definition of relation degrees in genealogy (i.e., consanguinity) is different from the pure graph-theory mathematical definition implemented in BFS [12]. For example, parents are considered first-degree relations in genealogy (based on the ontology), while they are considered to be second-degree relations mathematically, since there is a family node between the parent and the child (i.e., the parent and the child are not connected directly), with siblings considered to be second-degree relations in both genealogy and graph theory. Combined BFS-DFS algorithms such as Random Walks [39] do not take into account domain knowledge and sample nodes randomly. In the genealogical research field, several traversal algorithms have been suggested for user interface optimization [56]. However, these algorithms aim to improve interfaces and user experience and are not suitable for complete data extraction (graph to text) tasks.

This paper presents a new traversal algorithm, Gen-BFS, which is essentially the BFS algorithm adapted to the genealogical domain. The Gen-BFS algorithm is formally defined as follows:

Algorithm 1

Gen-BFS algorithm.

**Input:** Node (*SP*), Depth (*D*)
**Output:** Traverse queue (*TQ*)
**Initialization:** Node queue (*NQ*), Depth queue (*DQ*), Current depth (*CD* = 0), Nodes to depth increase (*NTDI* = 1), Next nodes to depth increase (*NNTDI* = 0)

1.    *NQ* **enqueue** *SP*
2.    **while** *NQ* **is not empty**
3.       n = *NQ* **dequeue**
4.       *DQ* **enqueue** n
5.       **if** n **is** Person
6.          kn = n→{$fam_{child}$} **union** n→{$fam_{parent}$}
7.       **else**
8.          kn = n→{$child_{fam}$} **union** n→{$parent_{fam}$}
9.       *NNTDI* = *NNTDI* + **count** (kn **not in** *NQ*)
10.      *NTDI* = *NTDI* – 1
11.      **if** *NTDI* = 0
12.         **if** n **is** Person
13.            *CD* = *CD* + 1
14.            **if** *CD* > *D*
15.               **break while**
16.         *NTDI* = *NNTDI*
17.         *NNTDI* = 0
18.      **for** n **in** kn
19.         **if** n **not in** *NQ*
20.            *NQ* **enqueue** n
21.    **while** *DQ* **is not empty**
22.       dn = *DQ* **dequeue**
23.       *TQ* **enqueue** dn
24.       **if** dn **is** Person
25.          **for** f **in** dn→{$fam_{parent}$}
26.             **for** p **in** f→{$parent_{fam}$}
27.                **if** p **not in** *DQ* **and** p **not in** *TQ*
28.                   *TQ* **enqueue** p
29.    **return** *TQ*

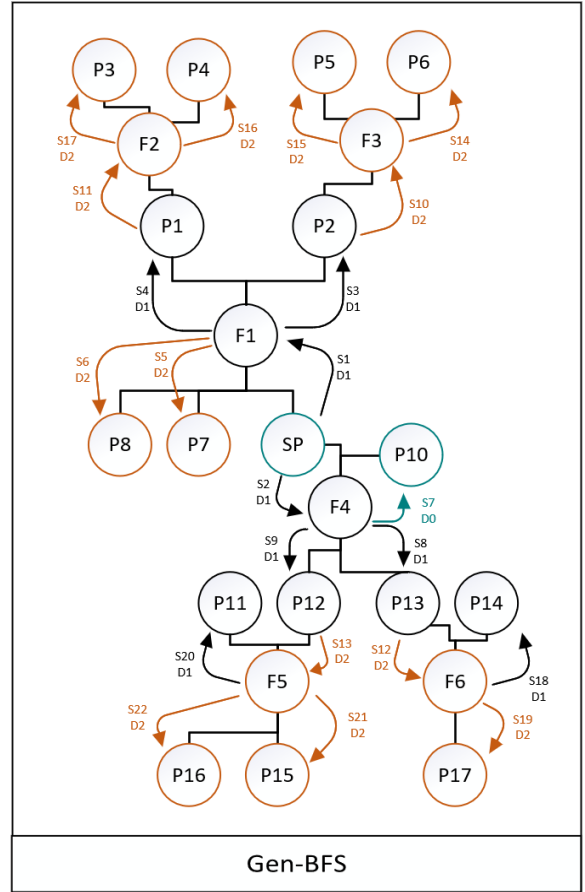

Fig. 8. Gen-BFS algorithm[15].

Where each node can be a Person or a Family, each Person node has two links (edges) types: $fam_{child}$ (FAMC in GEDCOM standard) and $fam_{parent}$ (FAMS in GEDCOM standard), each Family has the opposite edge types: $child_{fam}$ and $parent_{fam}$. Where {$fam_{child}$} is the collection of all the families in which a person is considered a child (biological family and adopted families), {$fam_{parent}$} is the collection of all the families in which a person is a parent (spouse) (i.e., all the person's marriages), {$child_{fam}$} is a collection of all the persons that are considered to be children in a family and {$parent_{fam}$} is a collection of all the persons considered to be a parent in a family. For example, the SP in Figure 2 is linked to two nodes. The link type to F1 is $fam_{child}$, and the link type to F4 is $fam_{parent}$. The family F1 in Figure 2 has two types of links. The link type to SP, P7, P8 is $child_{fam}$, and the link type to P1 and P2 is $parent_{fam}$.

---

[15] An algorithm step is noted as S. The degree of relation is noted as D. Relations are color-coded as follows: Zero-degree relation (self) - turquoise, First-degree relations – black, and Second-degree relations – brown.

Figure 8 illustrates the Gen-BFS traversal applied to the family tree presented in Figure 2. As shown in Figure 8, Gen-BFS is aware of the genealogical meaning of the nodes and reduces the tree traversal's logical depth. It ignores families in terms of relation degree, considers SP's spouses as the same degree as SP and SP's parents and children as first degree, and keeps siblings and grandparents as second-degree. In particular, lines 1-20 in Algorithm 1 represent a BFS-style traverse over the graph. In lines 5-8, the algorithm introduces domain knowledge and adds nodes to its queue according to the node type. The code in lines 9-17 ensures that the traversal will stop at the desired depth level. If the current node is a Person (line 12) and the current depth (CD) is about to get deeper than the required depth (D), then the while loop will end (line 14). Otherwise, the Persons and Families in the current depth (kn) will be added to the node queue (NQ) and may (depending on the stop mechanism) be added to the depth queue (DQ). In line 21, the depth queue (DQ) holds all the Family nodes and most of the Person nodes (except for spouses of the last depth level's Person nodes) within the desired depth level. For example, traversing with D = 1 over the family tree in Figure 2 will result in DQ that contains SP and her children and parents (F1, F4, P10, P1, P2, P12, and P13). However, according to the genealogical definition of depth levels in a family relationship, the children's spouses, P11 and P14 (but not the grandchildren, F5 and F6, which belong to D = 2) should also be retrieved. Lines 21-28 address this issue and add the missing Person nodes, thus logically reducing the depth of the graph.

Each family tree was split into sub-graphs using the Gen-BFS algorithm. New sub-graphs were created for each person as SP (source person) and its relations at different depth levels. Therefore, there is an overlap between the sub-graphs (a person can appear in several sub-graphs), and the sub-graphs cover all the individuals and relations in a given family tree. The Gen-BFS traversal algorithm is used both for dataset generation and for selecting the scope of the user's query in the inference phase (i.e., when answering the question).

Once extracted, each genealogical sub-graph was presented as a knowledge graph. This study adopted an event-based approach to data modeling presented in the past literature ([2, 31, 113]). As in [113], a formal representation of the GEDCOM heterogeneous graph (excluding the unstructured texts) as a knowledge graph was implemented using CIDOC-CRM, but in a more specific manner (e.g., we used concrete events and properties such as *birth, brought into life* as opposed to [113] that used generic vocabulary). We chose to use CIDOC-CRM as it is a living standard (ISO 21127:2014) for cultural heritage knowledge representation. CIDOC-CRM is designed as "a common language for domain experts" and "allows for the integration of data from multiple sources in a software and schema-agnostic fashion" [60]. It has been applied as a base model and extended in many domains related to cultural heritage, and in this study, it was chosen as a basis for defining the genealogical domain ontology due to its standard and generic nature and event-based structure, that enables *n*-ary rather than binary relationships between entities in the ontology, as required for representing genealogical and biographic data based on events in families and person's lives (e.g., E67 represents a birth event that connects a person, a place and a time span).

Genealogical graphs contain instances of two explicit classes: Person (E21 in CIDOC-CRM) and family that can be represented as a Group (E74 in CIDOC-CRM); and several implicit classes: Place (E53), Event (E5), Death (E69), Birth (E67) and others. These implicit classes are not structured as separate entities in the GEDCOM standard, but need to be extracted from the GEDCOM attributes. Properties matching various GEDCOM relations can also be easily found in CIDOC-CRM, e.g., the relation of a person to its children can be represented using P152 (is parent of).

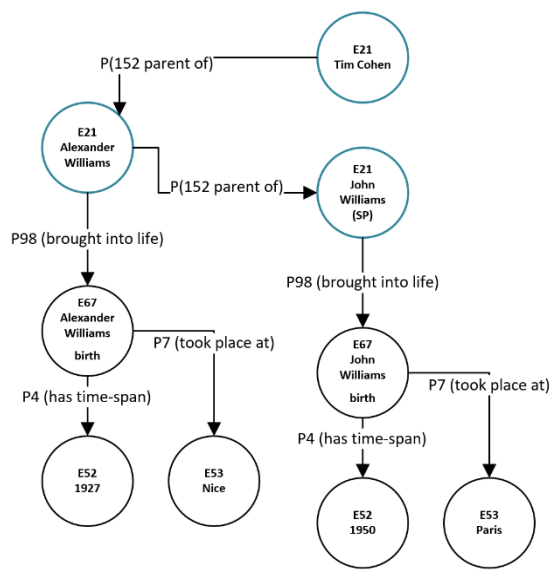

Fig. 9. GEDCOM individual's knowledge graph in the CIDOC-CRM-based format.

Figure 9 is an example of a representation of the GEDCOM sub-graph as a knowledge graph. As illustrated in the figure, the SP node is an instance of the class Person and has a relation (property) to a birth event (E21=>P98=>E67) with a relation to the place, Paris (E67=>P7=>E53) and a relation to the birth year with the value 1950 (E67=>P4=>E52). Representing GEDCOM as a knowledge graph is a critical step as the dataset generation method is based on well-established knowledge-graph algorithms, as described next.

*3.1.2. Text passage generation*

Next, a textual passage from each sub-graph is generated, representing the *SP's* genealogical data based on the graph-to-sequence. Text passages were generated using a knowledge-graph-to-text DNN model [68] and completed (for low model confidence or missing facts) with knowledge-graph-to-text template-based methodology [76]. It is important to note that converting the obtained genealogical knowledge sub-graphs to text is a more straightforward task than the open domain knowledge-graph-to-text or generic commonsense knowledge-graph-to-text task, since they are well structured and relatively limited in their semantics. For example, the sub-graph presented in Figure 9 can be converted to a sentence with template rules or using DNN models. A rule example will be: [First Name] [Last Name] *was born in* [Birth Year] *in* [Birthplace] = "John Williams was born in 1950 in Paris".

Using a knowledge-graph-to-text DNN model [68] and a knowledge-graph-to-text templates methodology [76], multiple variations of sentences conveying the same facts (comprised of the same nodes and edges in the graph) were composed based on different templates and combined with the sentence paraphrasing using a DNN-based model (the model of [63]). Most of the text passages were generated using a DNN model. However, the template-based method added variations that the DNN model did not capture. Table 1 above presents examples of such sentences created for the sub-graph in Figure 9.

Another critical challenge resolved by this approach is the multi-hop question answering problem, where the model needs to combine information from several sentences to answer the question. Although there are multi-hop question answering models presented in the literature [30, 74], their accuracy is significantly lower than a single-hop question answering.

To illustrate the problem, consider a user asking about the SP's (John's) grandfather: "Where was John's grandfather born?" or "Where was Tim Cohen born?", where Tim Cohen refers to John's grandfather. To answer both questions without multi-hop reasoning for resolution of multiple references to the same person, the graph-to-text template-based rules include patterns that encapsulate both the SP's relationship type (John's grandfather) and the relative's name (Tim Cohen), thus allowing the model to learn that Tim Cohen is John's grandfather. There are three types of references to a person that allows the DNN model to resolve single or multi-hop questions: 1) Direct referencing to a person with his/hers first and last name (e.g., John Williams), 2) Partial referencing to a person with his/hers first or last name (e.g., John), and 3) Multi-hop encapsulation, i.e., referencing to a person with their relative name to the SP (e.g., Alexander's son).

As a result of the above processing, multiple text passages were created for each SP's sub-graph. Since each sentence is standalone and contains one fact, sentences were randomly ordered within each text passage. Thus, even if the passage is longer than the neural model's computing capability, the model will likely encounter all types of sentences during its training process. These text passages were further encoded as vectors (i.e., embeddings) to train a DNN model that learns contextual embeddings to predict the answer (i.e., start and end positions in the text passage) for a given question.

Table 1

Genealogical-knowledge-graph-to-text context template example.

| Template-based rule example | Result | Reference type |
| --- | --- | --- |
| [First Name] [Last Name] was born in [Birth Year] in [Birthplace] | John Williams was born in 1950 in Paris | Direct |
| [First Name] was born in [Birth Year] in [Birthplace] | John was born in 1950 in Paris | Partial |
| [Name relative of SP] ([First Name] [Last Name]) was born in [Birth Year] in [Birthplace] | Alexander's son (John Williams) was born in 1950 in Paris | Multi-hop encapsulation |
| [First Name] was born in [Birthplace] in [Birth Year] | John was born in Paris in 1950 | Partial |
| [Relative First Name] [Relative Last Name] ([Relation to SP]) was born in [Birth Year] in [Birthplace] | Alexander Williams (John's father) was born in 1927 in Nice. | Multi-hop encapsulation |
| In [Birth Year] [First Name] was born | In 1950 John was born | Partial |
| [Birthplace] was [First Name] 's birthplace | Paris was John's birthplace | Partial |

### 3.1.3. Generation of questions and answers

Using the generated text passages (contexts), pairs of questions and answers were created. The answers were generated first, and then the corresponding questions were built for them as follows. Knowledge graph nodes and properties (relationships), as well as named entities and other characteristic keywords extracted from free text passages were used as answers. To achieve extensive coverage, multiple approaches were used for generation of questions. First, a rule-based approach was applied for question generation from knowledge graphs [140] and a statistical question generation technique [44] was utilized for WH question generation from the unstructured texts in GEDCOM.

Most of the questions (73%) were created using these methods. To identify the types of questions typical of the genealogical domain and define rule-based templates for their automatic generation, this study examined the genealogical analysis tasks that users tend to perform on genealogical graphs [10]. These tasks include: (1) identifying the SP's ancestors (e.g., parents, grandparents) or descendants (e.g., children, grandchildren), (2) identifying the SP's extended family (second-degree relations), (3) identifying family events, such as marriages, (4) identifying influential individuals (e.g., by occupation, military rank, academic achievements, number of children), and (5) finding information about dates and places, such as the date of birth, and place of marriage [4, 10]. These analysis tasks were adopted to define characteristic templates for natural language questions that a user may ask about the SP or its relatives. Some of these questions can be answered directly from the structured knowledge graph (e.g., "When was Tim's father born?"), while others can only be answered using the unstructured texts attached to the nodes (e.g., "Did Tim's father have cancer?").

A DNN-based model for generating additional types of questions [25] was used to complement the rule-based method. The neural question generation model predicted questions from all the unstructured texts in the GEDCOM data and produced 24% of the questions in the dataset (excluding duplicate questions already created using the WH-based and rule-based approaches).

Table 2

Knowledge-graph-to-text question template examples.

| Template-based rule example | Result |
| --- | --- |
| How many children did [First Name] [Last Name] have? | How many children did John Williams have? |
| How many grandchildren did [Relative First Name] [Relative Last Name] ([Relation to SP]) have? | How many grandchildren did Alexander Williams (John's father) have? |
| Was [Birthplace] [First Name] 's birthplace? | Was Paris John's birthplace? |

Finally, additional rules were manually compiled using templates [1, 28] to create questions missed by previous methods, mainly quantitative and yes-no questions (as illustrated in Table 2). These questions were 3% of all the questions in the datasets. All answer indexes were tested automatically to ensure that the answer text exists in the context passage. A random sample of 120 questions was tested manually by the researchers as a quality control process, and the observed accuracy was virtually 100%. However, it is still possible that DNN generated some errors. Nevertheless, even in this case, the study's conclusions would not change, as such errors would have a similar effect (same embeddings) on all the tested models.

*3.2 Fine-tuning the BERT-based DNN model for question answering*

Fine-tuning a DNN model is the process of adapting a model that was trained on generic data to a specific task and domain [22]. An initial DNN model is usually designed and trained to perform generic tasks on large domain-agnostic texts, like Wikipedia. In the case of the open-domain question answering, the BERT baseline model was pre-trained on English Wikipedia and Books Corpus [139] using Masked Language Modeling (MLM) and Next Sentence Prediction (NSP) objectives [22]. The MLM methodology is a self-supervised dataset generation method. For each input sentence, one or more tokens (words) are masked, and the model's task is to generate the most likely substitute for each masked token. In this fill-in-the-blank task, the model uses the context words surrounding a mask token to try to predict what the masked word should be. The NSP methodology is also a self-supervised dataset generation method. The model gets a pair of sentences and predicts if the second sentence follows the first one in the dataset. MLM and NSP are effective ways to train language models without annotations as a basis for various supervised NLP tasks. Combining MLM and NSP training methods allow modeling languages with both word-level relations and sentence-level relations understanding. The pre-trained BERT-based question answering model was designed with 12 layers, 768 hidden nodes, 12 attention heads, and 110 million parameters. Using such a pre-trained model, DNN layers can be added to fit to a specific task [22].

As shown in Figure 10, a new BERT-based model, Uncle-BERT, was fine-tuned for genealogical question answering as follows: (1) adding a pair of output dense layers (vectors) with dimensions of the hidden states in the model (*S* and *E)*, (2) computing the probability that each token in these layers (vectors) is the start (*S*) or end (*E*) of the answer, and finally (3) running and tuning the baseline BERT model described above for learning *S* and *E*. The probability of a token being the start or the end of the answer is the dot product between the token's numerical representation (i.e., embeddings) in the last layer of BERT and the new output layers (vectors *S* or *E*), followed by a softmax activation over all the tokens. Then, using the genealogical training dataset, the model is trained to solve the task in the study's domain. It should be noted that generation methods for pre-trained static node embeddings like node2vec [39] or TransE [14] treat triples as the training instance for embeddings, which may be insufficient to model complex information transmission between nodes. Therefore, the information is encoded from graph nodes into syntactic sentences and then the original BERT approach [22] is applied to generate comprehensive contextual embeddings from these sentences [43].

Figure 11 summarizes the developed genealogical question answering pipeline. To simplify the task, the proposed architecture asks the user to first select the family tree from the corpus (future research can eliminate this step by embedding the family trees [37] and ranking them based on similarity to the question [92]). As demonstrated in the figure, the family tree corpus (comprised of GEDCOM files) is processed into question answering datasets for different scopes. The process starts when a user selects a specific person from a family tree. Then the user indicates a scope (a genealogical relation degree, as described in Figure 1) to ask about (e.g., the *SP* itself, first-degree relative, second-degree relatives) and asks a question ("What was Alexander's father's military rank?"). The Gen-BFS algorithm incorporates the SP and the scope to generate a text passage that encapsulates the *SP*'s scope aligned with the user intent (equivalent to finding the top K text passages in the open-domain question answering pipeline). Finally, a fine-tuned DNN model, selected based on the requested relational degree (i.e., a model trained to predict answers on the requested relational degree), predicts the answer using the generated text passage and a question as inputs.

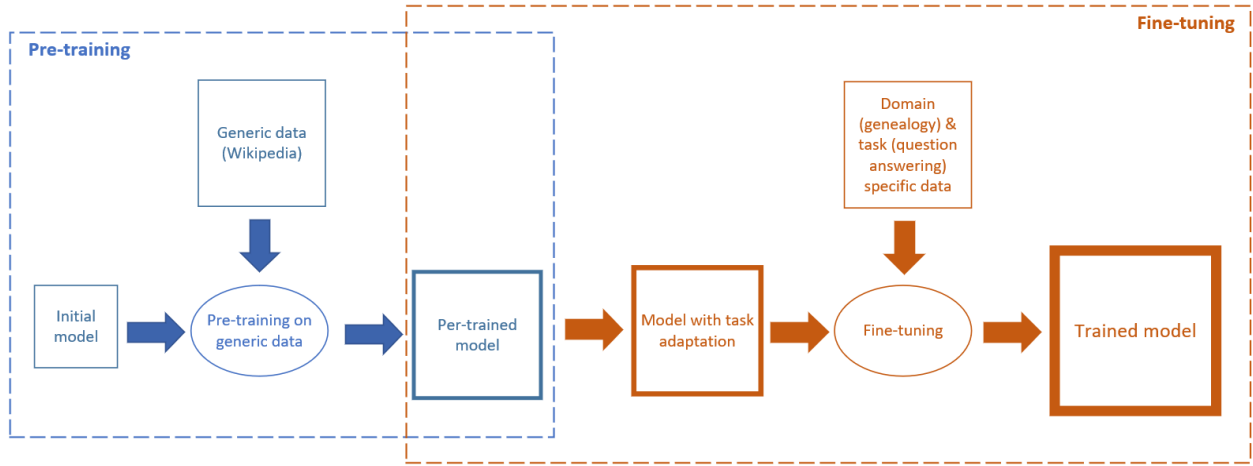

Fig. 10. The DNN model fine-tuning process.

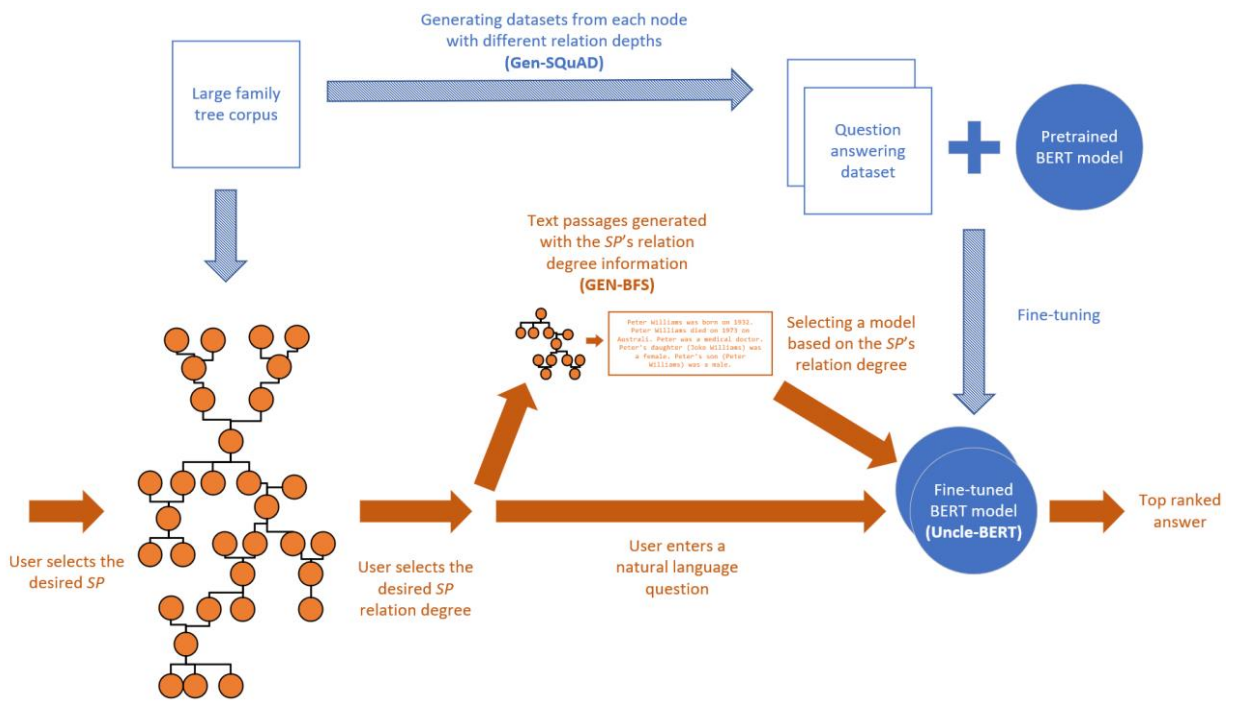

Fig. 11. Genealogical question answering pipeline (the proposed architecture).

## 4. Experimental design

This section describes the experimental dataset and training conducted to validate the proposed methodology for the genealogical domain.

*4.1. Datasets*

In this research, 3,140 family trees containing 1,847,224 different individuals from the corpus of the Douglas E. Goldman Jewish Genealogy Center in Anu Museum[16] were used. The Douglas E. Goldman Jewish Genealogy Center contains over 5 million individuals and over 30 million family tree connections (edges) to families, places, and multimedia items. To comply with the Israeli privacy regulation[17] and the European general data protection regulation[18] (GDPR), only family trees for which the Douglas E. Goldman Jewish Genealogy Center in Anu Museum has been granted consent or rights to publish online were used in the dataset generation. Moreover, as far as possible, all records containing living individuals have been removed from the dataset. Furthermore, all personal information and any information that can identify a specific person in this paper's examples, including the examples in the figures, have been altered to protect the individuals' privacy.

From the filtered GEDCOM files belonging to the above corpus, and after removing some files with parsing or encoding errors, three datasets were generated: Gen-SQuAD$_0$ using zero relation degree (SP and its spouses) with 6,283,082 questions, Gen-SQuAD$_1$ using first-degree relations with 28,778,947 questions, and Gen-SQuAD$_2$ using second-degree relations with 75,281,088 questions. Although all generated datasets contain millions of examples, only 131,072 randomly selected questions were used from each dataset when training the Uncle-BERT models. These were enough for the models to converge. Therefore, the size of the dataset did not impact the training results.

Each dataset was split into a training set (60%), a test set (20%), and an evaluation set (20%). To better evaluate the success of the different question answering models, the 131,072 questions in each dataset were classified into twelve types. Examples of questions and their classification types are shown in Table 3. Each question may refer to the *SP's* relationship type (e.g., Emily's grandson or by the direct name of the relative, e.g., Grace) and target one type of ontological entity as an answer (date, place, name, relationship type). Questions were classified into types based on the template, if generated using the template-based method (e.g., templates using place attributes were classified as "place", and date attributes as "date"), based on the WH question (e.g., When questions were classified as "date", and Where as "place"), if generated using the WH generation algorithm, or as general information / named entity, if generated by the DNN model. Therefore, the information / named entity may also include the other types of questions. It is important to note that these questions are semantically similar to the open-domain questions in SQuAD [90, 91] datasets.

---

[16] https://dbs.anumuseum.org.il/skn/en/c6/e18493701

[17] https://www.gov.il/BlobFolder/legal-info/data_security_regulation/en/PROTECTION%20OF%20PRIVACY%20REGULATIONS.pdf

[18] https://gdpr-info.eu/

Table 3
Question types.

| Question type / objective | Examples | Source |
|---|---|---|
| Name | What is Emily's full name? <br> What is Emily's last name? | Rule-based <br> Rule-based |
| Date | When was Emily born? <br> When did Emily get married? | Rule-based <br> Rule-based |
| Place | Where was Emily buried? <br> Where did Emily live? | Rule-based <br> Rule-based |
| Information / named entity | Who was Emily's first boyfriend? <br> Did Emily go to college? | DNN <br> DNN |
| First-degree relation | Who was Emily's son? <br> Who was Jonathan? | DNN, rule-based <br> DNN |
| Second-degree relation | How many sisters did Emily have? <br> How many brothers did Emily have? | Rule-based <br> Rule-based |
| First-degree date | When was Emily's husband born? <br> When was John born? | Rule-based <br> Rule-based |
| First-degree place | Where was Emily's father born? <br> Where was Alexander born? | Rule-based <br> Rule-based |
| First-degree information / named entity | What was Emily's father's academic degree? <br> What was Alexander's illness? | DNN <br> DNN |
| Second-degree date | When did Emily's sister die? <br> When did Yalma die? | Rule-based <br> Rule-based |
| Second-degree place | Where was Emily's grandson born? <br> Where was Grace born? | DNN <br> Rule-based |
| Second-degree information / named entity | What was Emily's grandfather's rank in the military? <br> Where was Tim's first internship as a lawyer? | DNN <br> DNN |

*4.2. Uncle-BERT fine-tuning*

For fine-tuning Uncle-BERT[19], the generated Gen-SQuAD training datasets were used. Each context in the Gen-SQuAD$_0$, Gen-SQuAD$_1$, and Gen-SQuAD$_2$ datasets was lowercased and tokenized using Word-Piece [132].

**[CLS]** When was Grace Williams born? **[SEP]** Mia's daughter (Emily Brown) was born in 1961 in Poland. Emily Brown had one husband. Mia Brown was buried in Poland. Mia's daughter (Grace Williams) was born in 1983 in Herzliya. Mia's daughter (Grace Williams) was a female. Mia Brown died in Poland. John Smith is Emily's husband. John Smith married Emily Brown in Tel Aviv. Emily Brown and Grace Williams were Mia's children…**[CLS]**

Fig. 12. Uncle-BERT model input example.

Figure 12 presents the model's input, where the [CLS] tag, which stands for classifier token, is the beginning of the input, followed by the first part of the input - the question. The [SEP] tag, which stands for a separator, separates the first part of the input (i.e., a question) and the second part – the context. [CLS] at the end indicates the end of the input.

To evaluate the effect of the depth of the consanguinity scope on the model's accuracy, an Uncle-BERT model was trained for each of the three datasets: Uncle-BERT$_0$ using Gen-SQuAD$_0$, Uncle-BERT$_1$ using Gen-SQuAD$_1$, and Uncle-BERT$_2$ using Gen-SQuAD$_2$. All models were trained with the same hyperparameters, that are shown in Table 4.

Table 4

Uncle-BERT training hyperparameters.

| Hyperparameters | Value |
|---|---|
| Max question tokens | 64 |
| Max sequence tokens | 512 |
| Max answer tokens | 30 |
| Doc stride | 128 |
| Batch size | 8 |
| Learning rate | 3e-5 |
| Train size | 131,072 |
| Epocs | 20 |

Max question tokens is the maximum number of tokens to process from the question input; if the question input length was greater than the Max question tokens, it was trimmed. Max sequence tokens are the maximum tokens to process from the combined context and question inputs.

If the cumulative context and question length was longer than the Max sequence tokens hyperparameter value, the context was split into shorter sub-texts using a sliding window technique; the Doc stride represents the sliding window overlap size. For example, consider the following hyperparameters' values: the max sequence tokens hyperparameter is 25, the doc stride hyperparameter is 6, and the following training example: "[CLS] When was Matt Adler's father born? [SEP] Matt's father (Noah Adler) was born in 1950 in London, England. Matt's father (Noah Adler) was a male. Matt's brother (Joanne Adler) was a male. Matt Adler was born in 1975 in London, England. Matt's mother (Carol) was born in 1950. [CLS]"; the question of the training example contains 7 tokens, and 18 tokens are left for the context. Therefore, the context will be split into three training examples: 1) "[CLS] When was Matt Adler's father born? [SEP] Matt's father (Noah Adler) was born in 1950 in London, England. Matt's father (Noah Adler) was a male [CLS]" (i.e., tokens 1 to 18), 2) "[CLS] When was Matt Adler's father born? [SEP] father (Noah Adler) was a male. Matt's brother (Joanne Adler) was a male. Matt Adler was born in [CLS]" (i.e., tokens 12 to 30), 3) "[CLS] When was Matt Adler's father born? [SEP] a male. Matt Adler was born in 1975 in London, England. Matt's mother (Carol) was born in 1950. [CLS]" (i.e., tokens 24 to 42). The model will be trained with the same question on the three new examples; if the answer span does not exist in an example, it is considered unanswerable.

Max answer tokens is the maximum number of tokens that a generated answer can contain. Train size is the number of examples used from the dataset during the training cycle.

As is customary with the SQuAD benchmark, an F1 score was calculated to evaluate Uncle-BERT models:

$$F1 = 2 * \frac{precision * recall}{precision + recall}$$

Precision equals the fraction of correct tokens out of the retrieved tokens (i.e., words that exist in both the predicted and the expected answer), and recall equals the fraction of the correct tokens in the retrieved (pre-dicted) answer out of the tokens in the expected answer. This metric allows measuring both exact and partial answers.

---

[19] A link to the code: https://github.com/omrivm/Uncle-BERT

## 5. Results

To evaluate the accuracy of the proposed fine-tuned models, the Gen-SQuAD$_2$ dataset was used to represent a real-world use-case in which a user is investigating her genealogical roots with the genealogical scope of two relation degrees (generations)[20]. To compare the model's accuracy for each type of answer, an F1 score was calculated to evaluate every Uncle-BERT model (i.e., Uncle-BERT$_0$ trained on Gen-SQuAD$_0$, Uncle-BERT$_1$ trained on Gen-SQuAD$_1$, and Uncle-BERT$_2$ trained on Gen-SQuAD$_2$). An overall accuracy evaluation of the three models was performed by calculating the F1 score for a mix of random questions of all types.

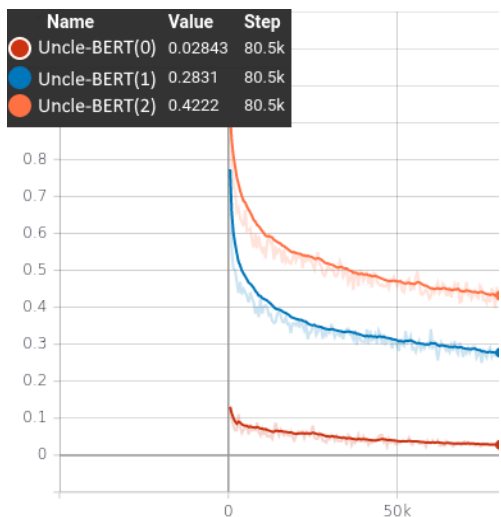

Fig. 13. The three Uncle-BERT model's train loss[21].

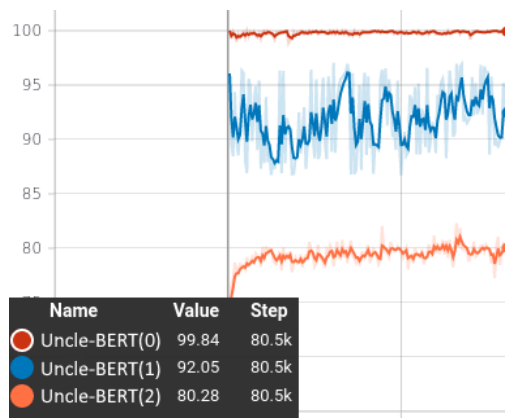

Fig. 14. The three Uncle-BERT model's train F1 score[22].

Figures 13 and 14 show the training loss and F1 scores of each of the three models. As expected, the more complex the context and questions, the lower the F1 score. While on narrow persons' contexts and questions (Gen-SQuAD$_0$), the model achieved an F1 score of 99.84; on second-degree genealogical relations (Gen-SQuAD$_2$), it achieved only an F1 score of 80.28.

Furthermore, as can be observed in Table 5, compared to the Uncle-BERT$_2$ model (trained with broader contexts of second-degree genealogical relations), the Uncle-BERT$_0$, which was trained using information about the *SP* and its spouses, fails to answer questions of any kind, including questions about the *SP* alone. We hypothesize that the model overfits to narrow contexts and therefore cannot handle larger context (Gen-SQuAD$_2$) "noise". This emphasizes the importance of the context size in the training data. Uncle-BERT$_1$ successfully answers most of the question types and even overtakes Uncle-BERT$_2$ in place-related questions. Except for place-related questions, it seems that a broader context improves the model's accuracy (Uncle-BERT$_2$).

Next, the best model, Uncle-BERT$_2$, was compared to several state-of-the-art open-domain question-answering DNN models. To this end, all the following models were trained using SQuAD 2.0 [90]: BERT [22], Distilbert [93], RoBERTa [70], Electra [18], DELFT [134]. Furthermore, to evaluate the effectiveness of the proposed genealogical question answering pipeline compared to the state-of-the-art knowledge graph-based pipeline, the genealogical adaptation of the DELFT model, Uncle-DELFT$_2$, was created. Uncle-DELFT$_2$, based on BERT combined with the GNN graph traversal, was trained on Gen-SQuAD$_2$.

---

[20] Similar to Anu Museum user interface - https://dbs.anumuseum.org.il/skn/en/c6/e8492037/Personalities/Weizmann_Chaim

[21] x-axis: number of epochs, y-axis: loss
[22] x-axis: number of epochs, y-axis: F1 score

As can be observed in Table 6, the baseline BERT model trained on the open-domain SQuAD 2.0 achieved an F1 score of 83 on the open-domain SQuAD 2.0 dataset [90]. However, on the genealogical domain dataset (Gen-SQuAD$_2$), it achieved a significantly lower F1 score (60.12) compared to the Uncle-BERT$_2$ (81.45). The fact that Uncle-BERT$_2$ achieves a higher F1 score is not surprising since the model was trained on genealogical data, as opposed to the baseline BERT model trained on the open-domain question data. However, when comparing Uncle-BERT$_2$ to Uncle-DELFT$_2$, it is clear that the performance improvement is due to the proposed methodology and not just due to the richer or domain-specific training data. Moreover, the DELFT method is much more complex than BERT, yet it achieved a lower score even when trained on the same domain-specific data. The fact that the vast majority of entities (found in both the "user" question and the expected answer) exists only in the unstructured data makes it hard for the GNN to find the correct answer (i.e., to complete the graph). This finding emphasizes the uniqueness of a genealogical question answering task compared to the open-domain question-answering and the need for the end-to-end pipeline and methodology for training and using DNNs for this task, as presented in this paper. Since Uncle-BERT$_2$ achieved a higher accuracy score than the more complex Uncle-DELFT$_2$ model, we conclude that the proposed method reduces complexity while increasing accuracy.

As shown in Table 6, although some questions appear in both Gen-SQuAD$_2$ and SQuAD 2.0 datasets, there is still a significant difference between open-domain questions and genealogical questions. Except for Uncle-DELFT$_2$ in the case of date questions, all the state-of-the-art models failed to answer natural genealogical questions compared to Uncle-BERT$_2$ (and in many cases, even compared to Uncle-BERT$_1$). However, Uncle-DELFT$_2$ was successful regarding date questions. This may imply that objective date questions are harder to extract from unstructured texts and the graph structure contributes to resolving such questions. Moreover, BERT's success on SP's date questions (compared to Uncle-BERT$_2$) may suggest that these questions are more generic and have more common features among different domains than unique features in the genealogical domain. Furthermore, the current state-of-the-art knowledge graph pipeline (i.e., DELFT) achieved performance similar to simpler BERT-based models. This indicates that while it is beneficial for open-domain questions, it is not as effective in the genealogical domain. This result, combined with the additional complexity of DELFT, makes it less satisfactory in this domain (except for date questions, as mentioned above).

Interestingly, the "basic" BERT model outperforms all the newer BERT-based models (except for Uncle-BERT$_2$). Furthermore, the fact that Uncle-BERT$_1$ achieved a higher F1 score on place type questions may indicate that place type questions may be more sensitive to "noise" or broad context. For example, place names may have different variations for the same entity (high "noise"), e.g., NY, NYC, New York, and New York City are all references to the same entity. This variety makes the model's task more difficult, thus adding broader contextual information and other types of "noise" (e.g., other entities, more people names, and dates), which may reduce the model's accuracy. Another possible reason for Uncle-BERT$_2$'s lower accuracy on place type questions may be the fact that Uncle-BERT$_2$ was trained with both one-hop-away and two-hop-away contexts while Uncle-BERT$_1$ was trained only with one-hop-away contexts. The fact that the F1 score of the model is smaller on second-degree place objective questions (1.39) than on first-degree (4.72) and zero-degree (10.01) place objective questions may reinforce this indication. However, it is important to notice that in many cases, this factor will not affect the F1 score since the F1 score does not use the position of the answer (start and end index), but only the selected tokens compared to the answer tokens. Since most children and parents live in the same place, either the parent's place (e.g., birthplace) or the child's place can be selected by the model without affecting the F1 score. Table 7 presents some examples of answer predictions for place objective questions by Uncle-BERT$_1$ and Uncle-BERT$_2$. These results suggest that higher accuracy can be achieved by classifying the question types and using a different model for different question types and relation depths.

Table 5

Uncle-BERT models F1 score on Gen-SQuAD$_2$.

| Question objective | Uncle-BERT$_0$ | Uncle-BERT$_1$ | Uncle-BERT$_2$ |
|---|---|---|---|
| Name | 44.53 | 95.14 | **97.64** |
| Date | 21.60 | 52.48 | **55.10** |
| Place | 27.54 | **88.53** | 78.52 |
| Information \ named entity | 16.91 | 15.22 | **87.40** |
| First-degree relation | 19.58 | 86.94 | **89.45** |
| Second-degree relation | 20.66 | 63.45 | **82.52** |
| First-degree date | 13.26 | 43.44 | **53.85** |
| First-degree place | 34.17 | **86.55** | 81.83 |
| First-degree information / named entity | 8.95 | 12.21 | **87.28** |
| Second-degree date | 11.68 | 43.12 | **44.87** |
| Second-degree place | 33.10 | **80.51** | 79.12 |
| Second-degree information / named entity | 8.37 | 11.34 | **81.04** |
| **Overall** | 19.73 | 69.92 | **81.45** |

Table 6

F1 scores of Uncle-BERT$_2$ and other state-of-the-art models on Gen-SQuAD$_2$.

| Question objective | BERT | Distilbert | RoBERTa | Electra | DELFT | Uncle-DELFT$_2$ | Uncle-BERT$_2$ |
|---|---|---|---|---|---|---|---|
| Name | 28.27 | 28.54 | 38.97 | 21.30 | 32.99 | 39.84 | **97.64** |
| Date | 60.58 | 53.30 | 44.33 | 34.92 | 39.62 | **79.35** | 55.10 |
| Place | 74.96 | 64.67 | 40.41 | 26.03 | 36.27 | 66.66 | **78.52** |
| Information / named entity | 71.20 | 66.79 | 34.96 | 31.72 | 40.91 | 70.58 | **87.40** |
| First-degree relation | 65.20 | 62.41 | 55.32 | 49.10 | 34.42 | 46.48 | **89.45** |
| Second-degree relation | 55.85 | 46.31 | 42.03 | 43.09 | 37.56 | 41.01 | **82.52** |
| First-degree date | 46.45 | 48.16 | 42.54 | 37.45 | 40.58 | **64.84** | 53.85 |
| First-degree place | 74.58 | 66.73 | 47.02 | 21.50 | 36.64 | 75.78 | **81.83** |
| First-degree information / named entity | 60.57 | 64.54 | 35.95 | 35.30 | 37.59 | 68.78 | **87.28** |
| Second-degree date | 39.49 | 39.75 | 23.30 | 26.99 | 38.29 | **60.15** | 44.87 |
| Second-degree place | 69.49 | 66.28 | 41.34 | 22.47 | 36.60 | 66.40 | **79.12** |
| Second-degree information / named entity | 60.19 | 62.38 | 34.37 | 37.15 | 35.70 | 47.26 | **81.04** |
| **Overall** | 60.12 | 60.19 | 39.45 | 43.39 | 37.56 | 42.96 | **81.45** |

Table 7

Uncle-BERT$_1$'s and Uncle-BERT$_2$'s prediction examples

| Question objective | Question | Context (relevant parts) | Correct Answer | Uncle-BERT$_1$ | Uncle-BERT$_2$ |
|---|---|---|---|---|---|
| Place | Where was John born? | … John was born in Poland in 1866 … John grew up in PL until he was… | Poland | in Poland | PL |
| Place | Where was John buried? | … John died and was buried in Germany during … Kate (John's daughter) was born in France… | Germany | Germany | France |
| First-degree place | Where did John's father get married? | … Matt (John's father) was born in Warsaw, Poland … Matt married Elain in Warsaw… | Warsaw | Warsaw | in Warsaw |
| First-degree place | Where was Matt killed? | … Matt died at home in Poland surrounded… his father (John's grandfather) was killed in Pruszkow in 1850… | Poland | Poland | Pruszkow |
| Second-degree place | Where did John's grandfather die? | … his father (John's grandfather) was killed in Pruszkow in 1850… | Pruszkow | in Pruszkow | Pruszkow |

## 6. Conclusions and future work

This study proposed and implemented a multi-phase end-to-end methodology for DNN-based answering natural questions using transformers in the genealogical domain.

The presented methodology was evaluated on a large corpus of 3,140 family trees comprised of 1,847,224 different persons. The evaluation results show that a fine-tuned Uncle-BERT$_2$ model, trained on the genealogical dataset with second degree relationships, outperformed all the open-domain state-of-the-art models. This finding indicates that the genealogy domain is distinctive and requires a dedicated training dataset and fine-tuned DNN model. A comparison of the proposed knowledge-graph-to-text approach was also found to be superior to the direct knowledge graph-based models, such as DELFT, even after domain-adaptation, both in terms of accuracy and complexity. This study also examined the effect of the type of question on the accuracy of the question answering model. The date-related questions are different as they can be answered with greater accuracy directly from the knowledge graph and may have more generic features than other question types, while place-related questions are more sensitive to noise than other question types. In addition, the evaluation results of the three Uncle-BERT models showed that the consanguinity scope of graph traversal used for generating a training corpus influences the accuracy of the models.

In summary, this paper's contributions are: (1) a genealogical knowledge graph representation of GEDCOM standard; (2) a dedicated graph traversal algorithm adapted to interpret the meaning of the relationships in the genealogical data (Gen-BFS); (3) an automatically generated SQuAD-style genealogical training dataset (Gen-SQuAD); (4) an end-to-end question answering pipeline for the genealogical domain; and (5) a fine-tuned question-answering BERT-based model for the genealogical domain (Uncle-BERT).

Although the proposed end-to-end methodology was implemented and validated for the question answering task, it can be applied to other NLP downstream tasks in the genealogical domain, such as entity extraction, text classification, and summarization. Researchers can utilize the study's results to reduce the time, cost, and complexity and to improve accuracy in the genealogical domain NLP research.

Possible directions for future research may include: (1) investigating the tradeoff between rich context passage generation and increasing the Gen-BFS scope, (2) integration with DNC or GNNs for dynamic scoping, (3) finding a method for classifying question types, (4) investigating the contribution of each question type to the accuracy of the model, and developing a model selection or multi-model method for each question type, (5) investigating larger contexts (relation degrees) using models that can handle larger input (e.g., Longformer [58] or Reformer [9]), (6) extending the Gen-BFS algorithm to handle missing family relations by adding a knowledge graph completion step while traversing the graph, (7) investigating the influence of the order of verbalized sentences and especially the order of person reference types, (8) investigating an architecture that will rank family trees (embedding the entire graph [37]) based on similarity to the question [92]) and eliminate the need for the user to select a family tree, (9) investigating the impact of spelling mistakes and out-of-vocabulary words on the quality of the results, (10) and training other transformer models on genealogical data to further optimize question answering DNN models for the genealogical domain.

with adapted pointer-generator networks. Neurocomputing, 382, 174-187. doi:10.1016/j.neucom.2019.11.079